\documentclass[10pt,journal,letterpaper,compsoc]{IEEEtran}
\usepackage{amsmath,graphicx}
\usepackage{dsfont}
\usepackage{booktabs,tabularx}

\ifCLASSOPTIONcompsoc
\else
\fi

\ifCLASSINFOpdf
\else
\fi

\begin{document}

\title{Efficient Region-Based Image Querying}

\author{S. Sadek, A. Al-Hamadi, B. Michaelis, U. Sayed \IEEEmembership{}

\IEEEcompsocitemizethanks{\IEEEcompsocthanksitem S. Sadek, A.
Al-Hamadi, and B. Michaelis, are with the Institute for
Electronics, Signal Processing and Communications (IESK),
Otto-von-Guericke-University Magdeburg, Magdeburg, Germany.\protect\\
E-mail: \{Samy.Bakheet, Ayoub.Al-Hamadi\}@ovgu.de.
\IEEEcompsocthanksitem U. Sayed is with Electrical Engineering
department, Assuit University, Assiut, Egypt.}}

\markboth{Journal Of Computing, Volume 2, Issue 6, June 2010,
ISSN 2151-9617 } {}
%

\IEEEpubid{\makebox[\columnwidth]{\hfill \textbf{\copyright~2010 Journal of Computing }}%
\hspace{\columnsep}\makebox[\columnwidth]{\textbf{
http://sites.google.com/site/journalofcomputing/}\hfill}}


\IEEEcompsoctitleabstractindextext{
\begin{abstract}
Retrieving images from large and varied repositories using
visual contents has been one of major research items, but a
challenging task in the image management community. In this
paper we present an efficient approach for region-based image
classification and retrieval using a fast multi-level neural
network model. The advantages of this neural model in image
classification and retrieval domain will be highlighted. The
proposed approach accomplishes its goal in three main steps.
First, with the help of a mean-shift based segmentation
algorithm, significant regions of the image are isolated.
Secondly, color and texture features of each region are
extracted by using color moments and 2D wavelets decomposition
technique. Thirdly the multi-level neural classifier is trained
in order to classify each region in a given image into one of
five predefined categories, i.e., "Sky", "Building",
"Sand$\backslash$Rock", "Grass" and "Water". Simulation results
show that the proposed method is promising in terms of
classification and retrieval accuracy results. These results
compare favorably with the best published results obtained by
other state-of-the-art image retrieval techniques.

\end{abstract}

}
\maketitle \IEEEdisplaynotcompsoctitleabstractindextext
\IEEEpeerreviewmaketitle
\section{Introduction}
\IEEEPARstart{W}ith the advent of high powerful digital imaging
hardware and software along with the accessibility of the
internet, databases of billions of images are now available and
constitute a dense sampling of the visual world. As a result,
efficient approaches to manage, index, and query such databases
are highly required. Classifying and querying image database
are frequently based on low-level image features such as color,
texture, and simple shape features. One simple way to query in
an image database is to create a textual description of all
images in the database and then employ the text-based
information retrieval methods to query these databases based on
the textual descriptions. Unfortunately, this way is not
feasible for two reasons. For one, annotating all images has to
be manually done and it is a very time-consuming task
particularly with large-scale databases.  Secondly, it is very
hard to find enough words conveying all contents of the images
in the database. Generally speaking, due to the subjectivity of
human perception and the rich content of images, no textual
description can be fully complete. Furthermore, The similarity
of images usually depends on the user and the context of the
query. For example, in querying a general-purpose image
database, a radiographic image might be sufficiently labeled as
"radiograph", while on the contrary this does not suffice
within a medical database comprising many varieties of
radiographs \cite{Smeulders00}.

Potential problems associated with conventional methods of
image indexing and querying have spurred a rapid rise in demand
for techniques for querying image databases on the basis of
automatically-derived features such as color, texture and
shape; a technology now generally referred to as Content-Based
Image Retrieval (CBIR). Following almost ten years of intensive
research, CBIR technology is now  moving out of the laboratory
and away from the closed experimental model into more realistic
settings, in the form of commercial products like Virage
\cite{Gupta96} and QBIC \cite{Flickner95}. However, the
technology is still immature, and lacks important usability
requirements that hinders its applicability. Additionally,
opinion will remain sharply divided over the usefulness of CBIR
in handling real-life queries in large and diverse image
collections in the absence of hard evidence on the
effectiveness CBIR techniques in practice \cite{Sutcliffe97}.

In this paper, a novel neural system for image retrieval is
presented. The system is based on an adaptive neural network
model called multi-level neural network. This model could
determine nonlinear relationship between different features in
images.  Results of the proposed system show that it is more
effective and efficient to retrieve visual-similar images for a
set of images with same conception can be retrieved.

The remainder of the paper is organized as follows. A brief
review of previous studies regarding our work is given in
Section 2. Section 3 highlights multi-level activation
functions used by the multi-level neural model. In section 4, a
fast segmentation technique based on a mean shift algorithm is
presented. Color moments and multi-level wavelet decomposition
are discussed in section 5. In section 6, the proposed image
classification and retrieval approach is introduced. Section 7
presents the simulation results of the proposed approach and
Section 8 closes the paper with some concluding remarks.
\begin{figure*}[~t]
\begin{center}
\includegraphics[width=.95\textwidth, height=1.8in]{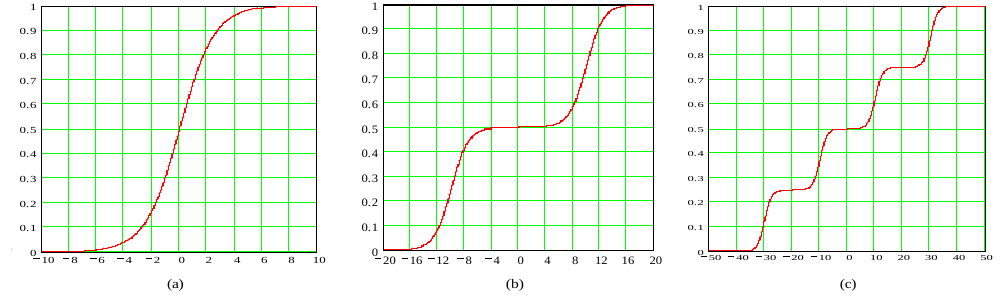}
\caption{Standard sigmoidal activation function and its
multi-level versions for $n=3$ and $n=5$.}\label{mlf}
\end{center}
\end{figure*}
\section{Related Work}
Over the course of the past two decade, a great deal of work
has been done to investigate and develop techniques for
classification and retrieval of large image databases
\cite{Deng04,Goodrum00,Connor05}. While the work of many
researchers within image classification and retrieval
\cite{Vailaya98,Zhao01,Luo01} has focused on more general,
consumer-level, semantic classes such as indoor/outdoor,
people/non-people, city, landscape, sunset, forest, etc , there
is another work done by other researchers that is centered on
the separation of computer-generated graphic images, such as
presentation slides, from photographic images
\cite{Hartmann02,Wang01,Prabhakar02}. In \cite{Stricker96}
authors use the histogram technique and refine the histogram to
include the absolute location of the pixel, or some kind of
homogeneity information into the histogram. These techniques
can work automatically as no preprocessing, like a
segmentation, is required. In \cite{Sadek09} Sadek et al.
propose a supervised method for classifying image contents into
four predefined categories. The classification process is done
without any pre-segmentation process. Furthermore, in
\cite{Sadek09c}, a system for content based image retrieval is
proposed. This system uses a neural model, called cubic spline
neural network. By using the spline neural model, the gap
between the low-level visual features and the high-level
concepts has been minimized. However image classification and
retrieval techniques are still far from achieving the goal of
being a complete, requiring additional research to improve
their effectiveness.

Machine learning in particular artificial neural networks is
increasingly employed to deal with many tasks of image
processing, e.g., image classification and retrieval. Amount
many classifiers, the neural classifier has the advantages of
being effortlessly trainable, highly rapid, and capable to
create arbitrary partitions of feature space \cite{Kuffler79}.
However a neural model, in the standard form, is incompetent to
correctly classify images into more than two categories
\cite{Bhattacharyya04}. This might be due to the fact that each
single processing element in this model, i.e. neuron, employs
standard bi-level activation function. As the bi-level
activation function only produces binary responses, the neurons
can generate only binary outputs. Therefore, in order to
produce multiple responses either an architectural or a
functional extension to the existing neural model is needed.

\section{Multi-level Neural Networks}
The pioneering work performed by McCulloch and Pitts
\cite{McCulloch49} in the area of Artificial Neural Networks
(ANNs) has initiated in 1943. Since then, there is an explosive
growth research in this field that has attracted and still
attracts many investigators in many disciplines such as
academician, physicians, psychologist, neurobiologist, etc. An
approach to the pattern recognition problem was introduced by
Rosenblatt \cite{Rosenblatt58} in his work on the perceptron.
Based on the literature, there are many successful projects and
on-going projects that are investigating the ability of neural
networks in their applications.	Theoretically, the applications
of neural networks are almost limitless but they can be
classified into several main categories such as classification,
modeling, forecasting and novelty detection. Some instances of
successful applications might include fault detection, credit
card, pattern recognition, handwritten character recognition,
color recognition, and share price prediction system, etc. Many
researchers have investigated the generalization capabilities
of the Artificial Neural Networks (ANNs) compared to the
traditional statistical methods such as Logistic Regression
(LR) models. The findings have revealed that neural networks
have a significantly better generalization capabilities than
those of other statistical methods such as regression
techniques \cite{Armoni98}.

As stated previously, a standard neural model employs bi-level
activation functions that produce only binary responses.
Instead, a multi-level neural model utilizes an activation
function, named a Multi-level Activation Function (MLAF). The
multi-level activation function originally is a functional
extension of the standard activation function. Several
multi-level forms belonging to several standard activation
functions can be defined.  We will show how to obtain the
multi-level form for an activation function from its standard
form. Let the standard sigmoidal activation function is given
by
\begin{equation}
f(x)=\frac{1}{1+e^{-\beta x}}
\end{equation}
where, $\beta$  is the steepness factor of the function. Thus
the multi-level form of the sigmoidal function can be derived
from the previous standard form as follows:
\begin{equation}
\varphi(x)\leftarrow f(x)+(\lambda-1)f(c)
\end{equation}
where $ (\lambda-1) c \leq x \leq \lambda c \textrm{ and }
1\leq \lambda\leq n.$ In Eq. (2), $\lambda$ represents the
color index, $n$ is the number of categories, and $c$
represents the color scale contribution. Fig. \ref{mlf} shows
the standard sigmoidal activation function and its
corresponding multi-level action functions for $n=3$ and $n=5$.
Note that the learning method of the MSNN model does not
considerably differ from the other learning methods used in
training artificial neural networks. It is employing some form
of gradient descent. This is done by taking the derivative of
the cost function with respect to the network parameters and
then altering those parameters in a gradient-related direction
\cite{Escudero00}.
\section{Image Segmentation}
\begin{figure}[~h]
\begin{center}
\includegraphics[width=1.0\columnwidth, height=2in]{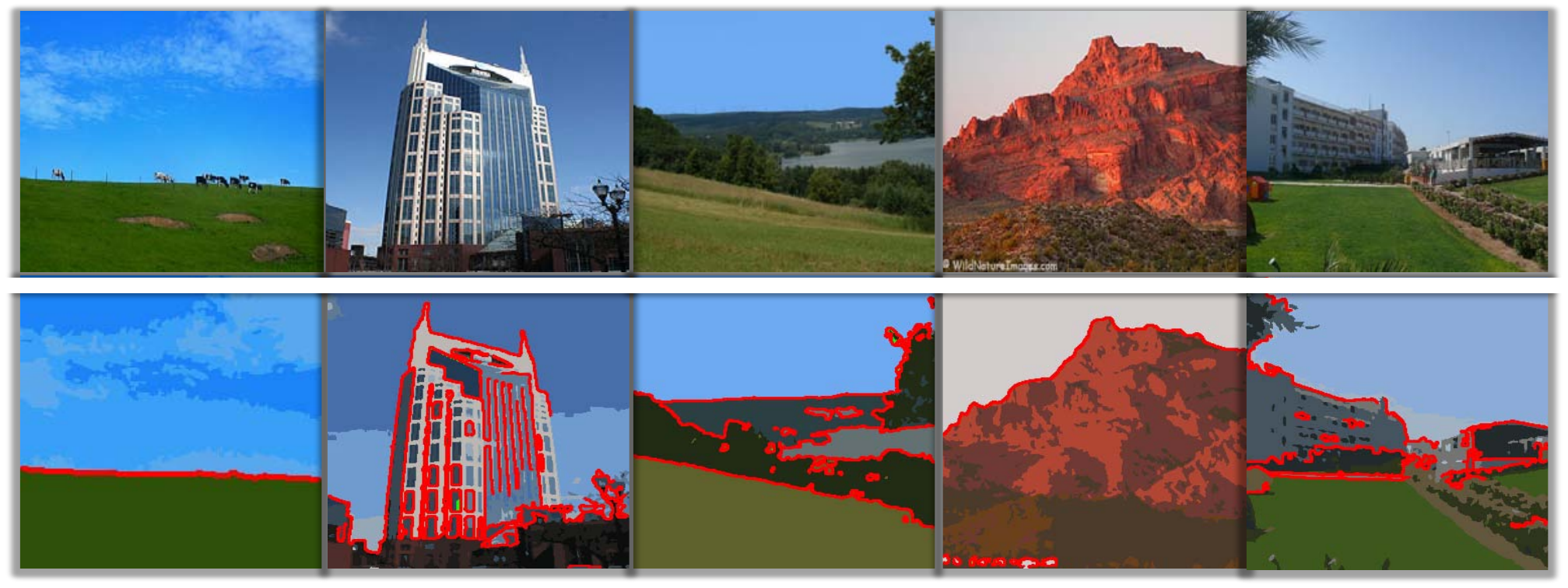}
\caption{Example of image segmentation based on the mean-shift
algorithm.}\label{fig1}
\end{center}
\end{figure}

In this section, a fast segmentation technique based on a mean
shift algorithm; a simple nonparametric procedure for
estimating density gradients is used to recover significant
image features (for more details see
\cite{Beveridge89,Cheng95}). Mean shift algorithm is really a
tool required for feature space analysis. We randomly choose an
initial location of the search window to allow the unimodality
condition to be settled down. The algorithm then converges to
the closest high density region. The steps of the color image
segmentation method are outlined as follows
\begin{enumerate}
  \item Initially, define the segmentation parameters (e.g.
      radius of the search window, smallest number of
      elements required for a significant color, and
      smallest number of contiguous pixels required for
      significant image regions).

  \item Map the image domain into the feature space.

  \item Define an appropriate number of search windows at
      random locations in the feature space.

 \item Apply the mean shift algorithm to each window to
     find the high density regions centers.

  \item Verify the  centers with image domain constraints
      to get the feature palette.

  \item Assign all the feature vectors to the feature
      palette using the information of image domain.

  \item Finally, remove small connected components of size
      less than a predefined threshold.
\end{enumerate}
It should be noticed that the preceding procedure is universal
and valid for applying with any feature space. Furthermore, all
feature space computations mentioned above are performed in HSV
space. An example of image segmentation by the previous mean
shift based algorithm is shown in Fig. \ref{fig1}.
\section{Feature Extraction} Image classification and
retrieval are regularly using some image features that
characterize the image. In the existing content-based image
classification and retrieval systems the most common features
are color, shape, and texture. Color histograms are commonly
used in image classification and retrieval. In this paper, we
use both color moments and approximation coefficients of
multi-level wavelet decomposition to extract features from each
image region.
\subsection{Wavelet Decomposition }
\begin{figure}[~h]
 \centering
\includegraphics[width=1.0\columnwidth]{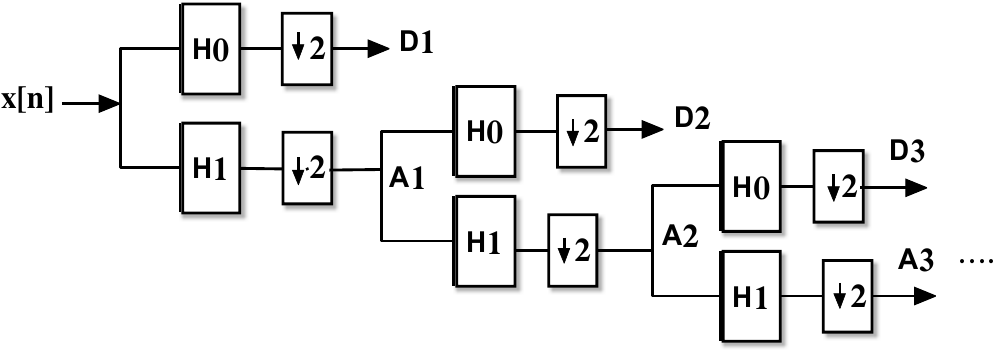}
 \caption{Multi-level wavelets decomposition.}
 \label{fig2}
\end{figure}

Discrete Wavelet Transform (DWT) captures image features and
localizes them in both time and frequency content accurately.
DWT employs two sets of functions called scaling functions and
wavelet functions, which are related to low-pass and high-pass
filters, respectively. The decomposition of the signal into the
different frequency bands is merely obtained by consecutive
high-pass and low-pass filtering of the time domain signal. The
procedure of multi-resolution decomposition of a signal x[n] is
schematically. Each stage of this scheme consists of two
digital filters and two down-samplers by 2. The first filter H0
is the discrete mother wavelet; high pass in nature, and the
second, H1 is its mirror version, low-pass in nature. The
down-sampled outputs of first high-pass and low-pass filters
provide the detail, D1 and the approximation, A1, respectively.
The first approximation, A1 is further decomposed and this
process is continued as shown in Fig. \ref{fig2}.
\begin{figure*}[~t]
  \begin{center}
  \includegraphics[width=.8\textwidth,height=1.8in]{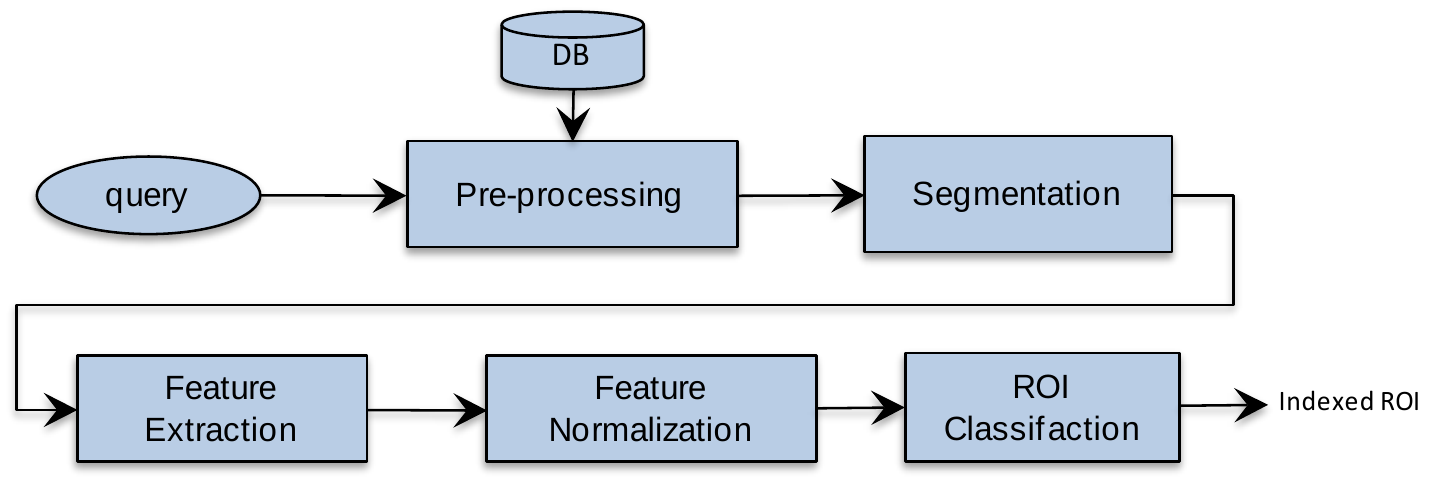}
    \caption{Block diagram of the proposed region-based image querying technique.}
  \label{outline}
  \end{center}
\end{figure*}
\subsection{Color Moments}
The basis of color moments lays in the assumption that the
distribution of color in an image can be interpreted as a
probability distribution \cite{Yu02}. Probability distributions
are characterized by a number of unique moments (e.g. normal
distributions are differentiated by their mean and variance).
It therefore follows that if the color in an image follows a
certain probability distribution, the moments of that
distribution can then be used as features to identify that
image based on color. The three central moments (Mean, Standard
deviation, and Skewness) of an image's color distribution can
be defined as
\begin{eqnarray}
  \mu_{k} &=& \frac{1}{n}\sum_{i=1}^{n}p_{i}^{k} \\
  \sigma_{k} &=& \sqrt{\frac{1}{n}\sum_{i=1}^{n}(p_{i}^{k}-\mu_{k})^{2}}\\
  s_{k} &=&\sqrt[3]{\frac{1}{n}\sum_{i=1}^{n}(p_{i}^{k}-\mu_{k})^{3}}
\end{eqnarray}
where $p_{i}^{k}$ is the value of the k'th color channel for
the i'th pixel, and $n$ is the size of the image.

\section{Proposed Approach}
The prime difficulty with any image retrieval process is that
the unit of information in image is the pixel and each pixel
has properties of position and color value; however, by itself,
the knowledge of the position and value of a particular pixel
should generally convey all information related to the image
contents \cite{Li08, Marmo05}. To surmount this difficulty,
features are extracted using two-way. The extracted features
consist of two folds: color moments and approximate
coefficients of multi-level wavelet decomposition. This allows
us to extract from an image a set of numerical features,
expressed as coded characteristics of the selected object, and
used to differentiate one class of objects from another. The
main steps of the proposed approach are depicted in Fig.
\ref{outline}. In the following subsections, the main steps of
the proposed approach depicted in Fig. 3 are described.
\subsection{Preprocessing }
In image processing, preprocessing mainly purpose to enhance
the image in ways that raise the opportunity for success of the
other succeeding processes (i.e. segmentation, features
extraction, classification,  etc). Preprocessing
characteristically deals with techniques for enhancing
contrast, segregating regions, and eliminating or suppressing
noise. Preprocessing herein includes normalizing the images  by
bringing them to a common resolution, performing histogram
equalization and applying the Gaussian filter to remove small
distortions without reducing the sharpness of the image.
\subsection{Segmentation }
In this step the fast mean shift based segmentation technique
 described above in section 2 is used to segment the image into
 distinct regions. To get rid of the segmentation errors,
regions of small area (i.e., less than a predefined threshold,
e.g., $t=0.05$) are discarded. The significant regions (i.e.
regions of areas greater than or equal 0.05 of the image area)
are the candidates where the feature vectors are extracted
from.
\subsection{Feature Extraction}
In this step, we utilize 2D multi-level wavelets transform to
decompose image regions. Each level of decomposition gives two
categories of coefficients, i.e., approximate coefficients and
details coefficients. Both approximate coefficients and color
moments are considered as the features for our retrieval
problem.
\subsection{ Feature Normalization }
To prevent singular features from dominating the others and to
obtain comparable value ranges, we do feature normalization by
transforming the feature component, $x$ to a random variable
with zero mean and one variance as follows
\begin{equation}\label{eq4}
   \bar{x}=\frac{x-\mu}{\sigma}
\end{equation}
where $\mu$ and $\sigma$ are the  mean and the  standard
deviation of the sample respectively. Suppose that each feature
is normally distributed, then the probability of $\bar{x}$
belonging in the [-1,1] range is 0.68. A further shift and
rescaling such as
\begin{equation}\label{eq:3}
\bar{x}=\frac{\frac{x-\mu}{3\sigma}+1}{2}
\end{equation}
would ensure that 0.99 of $\bar{x}$  values laying in [0,1].
\subsection{ Classification of Image Regions}
As a matter of fact, it should be stated that the  neural
classifier can accomplish better classification if each  region
belongs to only one of the predefined categories. Therefore, it
is hard to build up a full trustworthy classifier due to the
truth that different categories may have similar visual
features (such as Water and Sky categories). Before doing any
classification process, categories that reflect the semantics
in the image regions are first defined. Then multi-level neural
classifier has to learn the semantics of each category via the
"training" process. So it is possible now to classify a
specific region into one of the predefined semantic categories
which humans easily understand. To do so, extracted features of
the region are fed into the trained multi-level classifier and
then it directly predicts the category of that region.
\section{ Experimental Results}
In this section classification and retrieval results of the
proposed approach are presented. First, to train the
multi-level classifier, we have manually prepared a training
set comprising of 200 regions; on the average, 40 training
samples per category. o verify the ability of the proposed
approach in image classification and retrieval, we have used a
test set containing about 500 regions covering 5 categories,
"Sky", "Building", "Sand $\backslash$Rock", "Grass" and "Water.
Table 1 tabulates  the classification results done by the
proposed approach.


\begin{table}[h]
  \centering
  \caption{Region classification results over the dataset collections}
  \label{t1}
  \begin{tabular}{p{1.5in} p{1.5in}}
\toprule
    Category & Precision \\
   \hline
    Sky &  96\% \\
    Building & 91\%\\
    Sand$\backslash$Rock & 89\% \\
    Water & 98\% \\
    Grass & 95\% \\
    \hline
    Average & 93.8\% \\
   \bottomrule
  \end{tabular}
\end{table}
It should be noted that the raw figures tabulated in table 1
are considered as a quite quantitative measure for the
performance of the proposed approach, indicating to the high
performance of the proposed approach compared with other
classification approaches, specifically that has been proposed
in \cite{Ohashi03}.

\begin{figure}[~h]
\centering
\includegraphics[width=1.0\columnwidth]{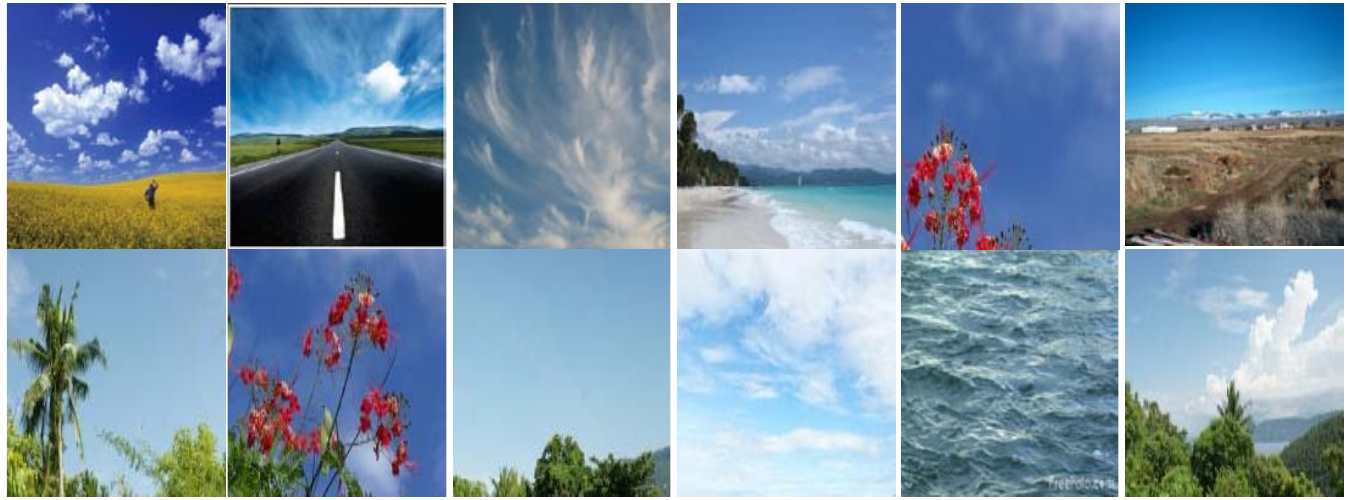}
\caption{Result of retrieval for the query: keyword
="sky".}\label{fig4}
\end{figure}

Once the semantic classification of image regions is
successfully done, in this case, an image  can be represented
by the categories in which the image regions are classified.
That is, each image can be characterized by a set of keywords
(i.e., categories' indices) which allows for things such as a
highly intuitive query to be possible. Therefore, by using one
or more keywords, image databases can easily be searched. In a
query of such type, all images in the database that contain the
selected keywords will be retrieved. For instance, if the
keyword "Sky" is selected. The purpose of this query is to
retrieve all images that include a region of sky. The retrieval
results of such query are shown in Fig. \ref{fig4}.
\section{Conclusion and Future Work}
In this paper, an efficient method for region-based image
classification and querying has been introduced. The method
employs a new classifier model, called multi-level neural
network. The low computational complexity as well as the
easiness of implementation are the key advantages of this
classifier model. The simulation results on image
classification and retrieval reveal that the multi-level neural
classifier is very effective in terms of learning capabilities
and retrieval accuracies. This allowed the method to give
promising retrieval results that compare favorably with those
obtained by other state-of-the-art image retrieval methods.
Although the current implementation of the method is tested on
a simple still image dataset, it can be easily extended and
applied on realistic video datasets. Such an issue is important
and will be in the scope of our future work.

\section*{Acknowledgments}
This work is supported by Forschungspraemie (BMBF-F\"orderung,
FKZ: 03FPB00213) and Transregional Collaborative Research
Centre SFB/TRR 62 "Companion-Technology for Cognitive Technical
Systems" funded by the German Research Foundation (DFG).

%
\vspace{-.2in}
\begin{IEEEbiography}[{\includegraphics [width=1.0in,height=1.25in]{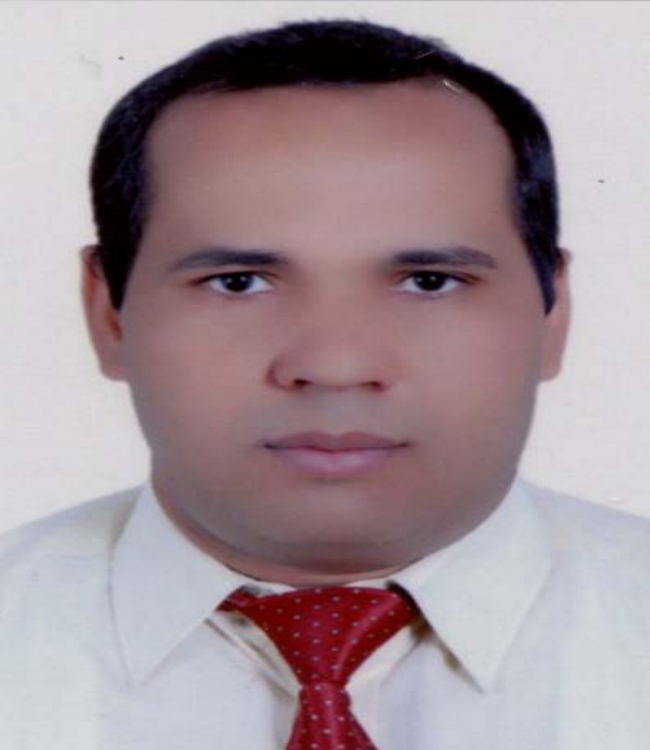}}]{Samy Sadek}
received the B.Sc. and M.Sc. degrees in Computer Science from
the University of Sohag, Sohag, Egypt in 2000, and 2005
respectively. He is currently working toward the Ph.D. degree
at the Institute of Electronics, Signal Processing, and
Communications (IESK), Otto-von-Guericke University Magdeburg,
Magdeburg, Germany. His current research interests include
Video retrieval, Human activity recognition, Visual
surveillance, Machine learning and Artificial Neural Networks
(ANN).
\end{IEEEbiography}
\vspace{-.8in}

\begin{IEEEbiography}[{\includegraphics [width=1.0in,height=1.2in]{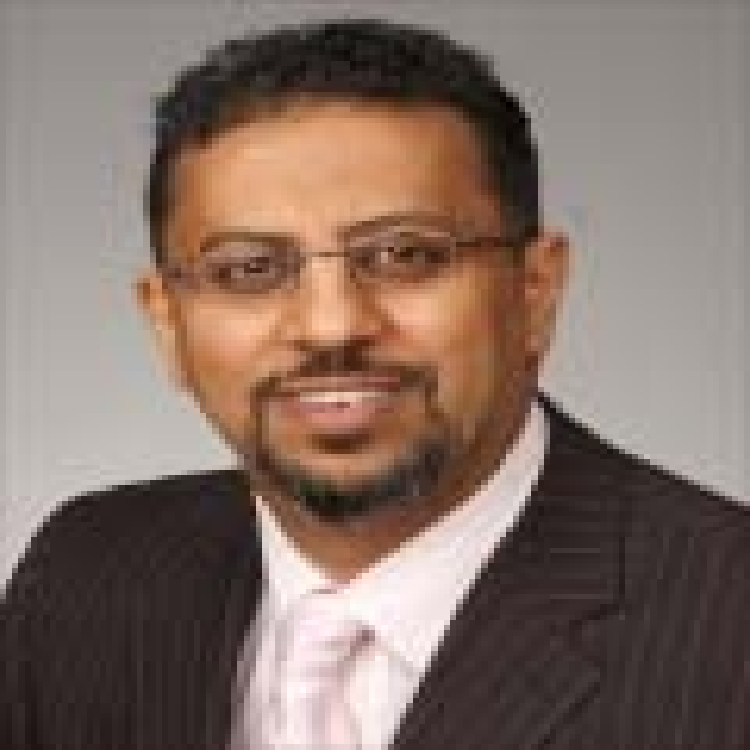}}]{Ayoub Al-Hamadi}
received his Masters Degree in Electrical Engineering and
Information Technology in 1997 and his PhD. in Technical
Computer Science at the Otto-von-Guericke-University of
Magdeburg, Germany in 2001. In May 2010 he received the
Habilitation in Artificial Intelligence and the Venia Legendi
in the scientific field of Pattern Recognition and Image
Processing. His research interests include Image processing,
Pattern recognition, Human Computer Interaction, Multi-objects
tracking, Video retrieval, Artificial Intelligence. Al-Hamadi
is the author of more than 120 articles in peer-reviewed
international journals, conferences and books.
\end{IEEEbiography}
\vspace{-.8in}
\begin{IEEEbiography}[{\includegraphics [width=1.0in,height=1.2in]{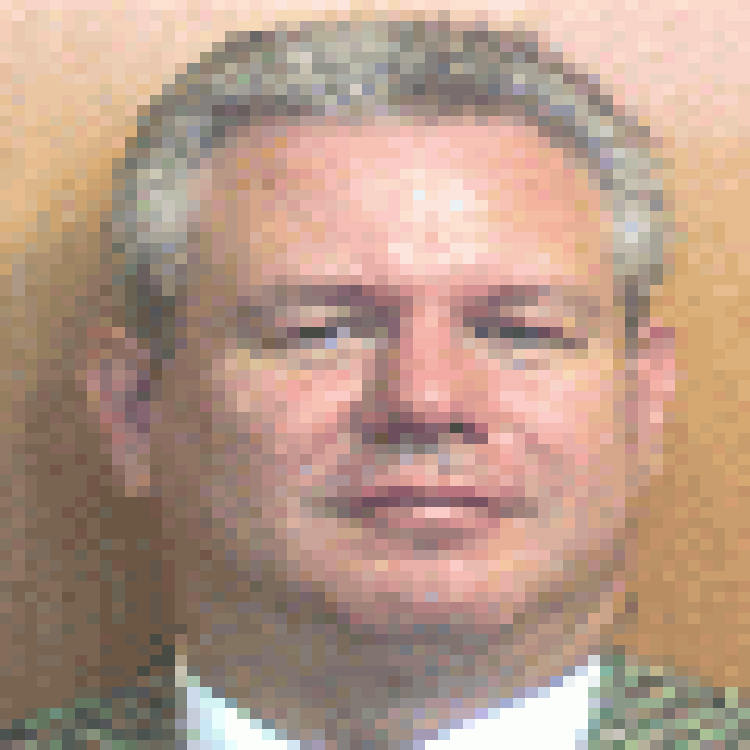}}]{Bernd Michaelis}
received the degree of Diploma Engineer for Electronics from
the Technische Hochschule Magdeburg in 1971. He received the
first and second doctoral degree in 1974 and 1980 respectively.
In 1980 he joined the Joint Institute for Nuclear Research
Dubna. In 1984 he became Hochschuldozent at the Technische
Hochschule Magdeburg. His research interests include Image
processing, Artificial neural networks, Biological neural
networks, Microcomputers and Processor architectures. Michaelis
is the author of more than 200 papers.
\end{IEEEbiography}
\vspace{-.8in}
\begin{IEEEbiography}[{\includegraphics [width=1.0in,height=1.2in]{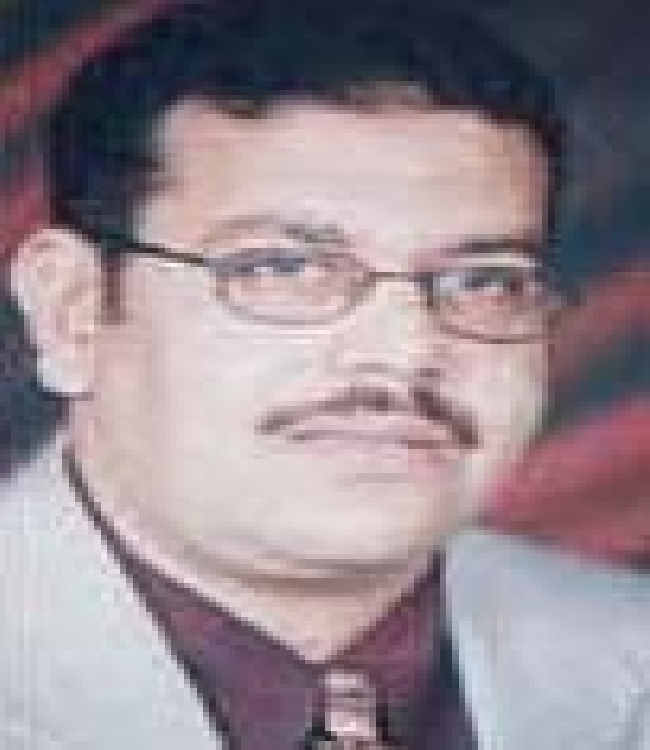}}]{Usama S. Mohammed }
received his B.Sc. and M. Sc. degrees from Assiut University,
in 1985 and 1993, respectively, and his Ph.D. degree from Czech
Technical University in Prague, Czech Republic, in 2000. From
November 2001 to April 2002, he was a post Doctoral Fellow with
the Faculty of Engineering, Czech Technical University in
Prague, Czech Republic. He authored and co-authored more than
55 scientific papers. His research interests include image
coding, speech coding, telecommunication technology,
statistical signal processing, blind signal separation,
wireless communication network, video coding, and RFID.
\end{IEEEbiography}
\end{document}